# Automated Question Generation for Science Tests in Arabic Language Using NLP Techniques

Mohammad Tami[1], Huthaifa I. Ashqar[2], Mohammed Elhenawy[3]

[1] Arab American University, Jenin, Palestine
[2] Arab American University Jenin, Palestine and Columbia University, NY, USA
Huthaifa.ashqar@aaup.edu
[3] CARRS-Q, Queensland University of Technology Brisbane, Australia

**Abstract.** Question generation for education assessments is a growing field within artificial intelligence applied to education. These question-generation tools have significant importance in the educational technology domain, such as intelligent tutoring systems and dialogue-based platforms. The automatic generation of assessment questions, which entail clear-cut answers, usually relies on syntactical and semantic indications within declarative sentences, which are then transformed into questions. Recent research has explored the generation of assessment educational questions in Arabic. The reported performance has been adversely affected by inherent errors, including sentence parsing inaccuracies, name entity recognition issues, and errors stemming from rule-based question transformation. Furthermore, the complexity of lengthy Arabic sentences has contributed to these challenges. This research presents an innovative Arabic question-generation system built upon a three-stage process: keywords and key phrases extraction, question generation, and subsequent ranking. The aim is to tackle the difficulties associated with automatically generating assessment questions in the Arabic language. The proposed approach and results show a precision of 83.50%, a recall of 78.68%, and an Fl score of 80.95%, indicating the framework's high efficiency. Human evaluation further confirmed the model's efficiency, receiving an average rating of 84%.

## 1 Introduction

In recent years, there have been significant improvements in the Natural Language Processing (NLP) field. One field that has been revolutionized due to these significant improvements is the field of education, by facilitating interactive and adaptive learning environments. Question generation from educational content is one of the main aspects that play a fundamental role in enhancing learners' comprehension, understanding, and knowledge. Arabic language is one of the most widely spoken languages globally. Due to Arabic language complexities, question-generation techniques in Arabic have emerged as a challenging area of research and development, its complex morphology, rich linguistic nuances, syntax, and semantics require special approaches for effective question generation.

The significance of the Arabic language cannot be overstated. It is not only utilized by hundreds of millions of individuals but also exhibits a rapidly expanding online presence, both in terms of users and content. Additionally, Arabic has several distinctive characteristics that present both a challenge and an intrigue in the automated processing and comprehension of Arabic text [1]. Despite the lack of available resources and their limitations, Arabic language processing (ANLP) over the last few years has gained more importance and several systems have been developed for a wide range of applications that deal with several complex problems due to the nature and structure of Arabic language [2]. After the significant advancement in the English language NLP and other languages, ANLP has received more attention, attracted many researchers, many ANLP research centers have been established, and several applications have been developed like sentiment analysis, spam detection, and text categorization [3].

Question generation task using the NLP technique, like any other data processing task, involves pre-processing that should be done before starting the question generation. In Preprocessing related to NLP, two types of preprocessing are usually performed: standard preprocessing and question generation (QG) specific preprocessing. Standard preprocessing is done by making tokenization, POS tagging, segmentation, sentence splitting, named entity recognition (NER), and relation extraction (RE), which are commonly used in NLP-related tasks. While QG-specific preprocessing focuses on the extraction of inputs that are more relevant to the GQ task. Three types of QG preprocessing are used: sentence simplification, sentence classification, and content selection [4].

In [5] [6] they used sentence simplification in some text-based approaches to simplify complex sentences with appositions or sentences joined with conjunctions to be used easily in upcoming NLP tasks. Relying on parse-tree dependencies, they simplify long Wikipedia sentences that contain multiple objects facilitating triplet extraction.

In [7] they used POS analysis and dependency labels to perform sentence classification to classify sentences to identify the type of question to be asked about the given sentence. Question generation requires identifying the most important contents, sentences, and concepts about which to generate the questions, and for this content selection, preprocessing is being used.

In [8] they outlined three criteria for the selection of important sentences for reading assessment questions construction, along they suggested metrics for evaluation: keyness, representing the core key meaning of the text; completeness, ensuring the comprehensive coverage of the text across various paragraphs; and independence, highlighting diverse aspects of the text content. Other researchers have adopted different approaches, selecting the most important sentences topically.

In [9] they selected sentences that reflected the main concepts and topics from an educational textbook. While In [10] [11], researchers rank sentences using topic modeling to identify topics based on topic distribution. [12] used another approach by mapping both the input document and its sentences into an identical n-dimensional vector space and then selecting sentences similar to the document within this space, under the assumption that such sentences most accurately represent the document's main topic or essence. Others [13] [14] filtered sentences that start with discourse connectives (e.g. thus, also, so, etc.) that are insufficient on their own to make valid questions.[15] They used (structured knowledge bases) DBpedia for content selection in conjunction with external ontologies that describe educational standards, these standards guided the selection of DBpedia content to generate questions.

On the other hand, pre-trained models (PTMs), or reusable NLP models for natural language processing (NLP), are deep learning models trained on large datasets and extensive corpora to perform various NLP tasks.PTMs already learned universal language representations useful for various downstream NLP tasks. By using (PTMs) to accomplish new tasks, we use the knowledge learned by the pre-trained model on a new dataset that may be different from the one used for training the pre-trained model; this saves significant time and resources compared to training a new model from scratch.

One notable contribution is the paper by Kumar, Banchs, and D'Haro titled "RevUP: Automatic Gap-Fill Question Generation from Educational Texts" [10], which addresses the challenge of automated gap-fill question generation in educational texts. The primary issues identified in the study include selecting relevant sentences, identifying appropriate gap phrases, and generating contextually fitting distractors. In terms of methodology, RevUP employs a three-part approach: Sentence Selection, Gap Selection, and Multiple Choice Distractor Selection. The Sentence Selection utilizes a novel ranking method based on topic distributions obtained from topic models. The Gap Selection involves a binary classifier trained on human annotations, achieving 81.0% accuracy. The study introduces a novel approach to choosing semantically similar distractors. The evaluation conducted through Amazon Mechanical Turk demonstrates that 94% of the selected distractors were considered good. The dataset used in this study includes sentences from Campbell's Biology textbook, with human annotations collected through Amazon Mechanical Turk contributing to the classifier's training for gap selection. The final dataset exhibits a slight skew towards bad gaps, with 554 bad gaps and 468 good gaps. RevUP's methodology presents advantages in filling the semantic gap left by previous methods, providing a significant advancement in automatically generating high-quality tests for teachers and self-motivated learners. The proposed methodology achieves a high accuracy of 81.0% in predicting the relevance of gap phrases. In our opinion, the study acknowledges limitations; the distractor selection task reflects low agreement (51%), suggesting the need for more precise evaluation with students/teachers. Additionally, the proposed method relies on topic models, and the impact of parameters on sentence pinpointing accuracy could be explored in future work.

The study in [16] explores the performance of the multilingual T5 model (mT5) [17] on Arabic language tasks and introduces new benchmarks (ARGEN) for both Arabic language understanding and generation. They compare mT5 with three newly pre-trained Arabic-specific text-to-text Transformer models and evaluate them on these benchmarks. Results show that the new Arabic language understanding and generation models outperform mT5 and exceed MAR-BERT (Multi-stage document ranking with BERT)[18], the current best Arabic BERT-based model. Due to their remarkable ability to transfer knowledge from unlabeled data to downstream tasks, pre-trained transformer-based language models have emerged as important components in modem natural language processing (NLP) systems. The pre-trained models utilized a learning rate of 0.01, a batch size of 128 sequences, and a maximum sequence length of 512, except for AraT5TW, which had a maximum sequence length of 128. The original T5 implementation in TensorFlow was employed for training over 1 million steps, taking approximately 80 days on a Google Cloud TPU with 8 cores (v3:8) from the TensorFlow Research Cloud (TFRC). The evaluation of their pre-trained language models involves the introduction of ARGEN, a novel benchmark designed specifically for Arabic language generation.

In Summary, several studies in the literature are showcasing innovative approaches to question generation. These include leveraging machine learning techniques, transformer models, template-based approaches, and the use of specific NLP tools to automate question generation from various text sources. These studies cover diverse applications, such as educational texts, multilingual models, mathematical problem generation, and automated test paper creation, demonstrating the versatility and potential impact of ANLP in various domains. They also address challenges and limitations in existing approaches, such as the need for larger annotated datasets, potential semantic comprehension limitations, and the necessity for more precise evaluation methods, especially in educational contexts. Previous research for Question generation requires the data to have a context/text and answers to be able to generate questions, our contribution in this paper is to overcome this limitation by building a framework that combines keyWord extracted (KeyBert) with a question generation pre-trained model (AraT5). Our framework receives the original text and the extracted keywords from keyBert and generates the question.

## 2 System Methodology

In this research, a comprehension methodology was implemented to ensure reliable and robust outcomes. Figure 1 outlines the procedural steps undertaken to conduct this research. The initial stage involved collecting data from actual Arabic textbooks, followed by data extraction. This extracted Arabic text served as the basis for creating keywords and key phrases that encapsulate the original material. These key phrases were then fed into a pre-trained transformer model (araT5), alongside the original text, to generate questions in Arabic for each passage. Subsequently, OpenAΓs GPT-4 was utilized to produce Arabic questions for the same content. These questions, generated by OpenAI, were used as a benchmark to evaluate our model, applying various metrics to conclude the model's effectiveness and efficiency.

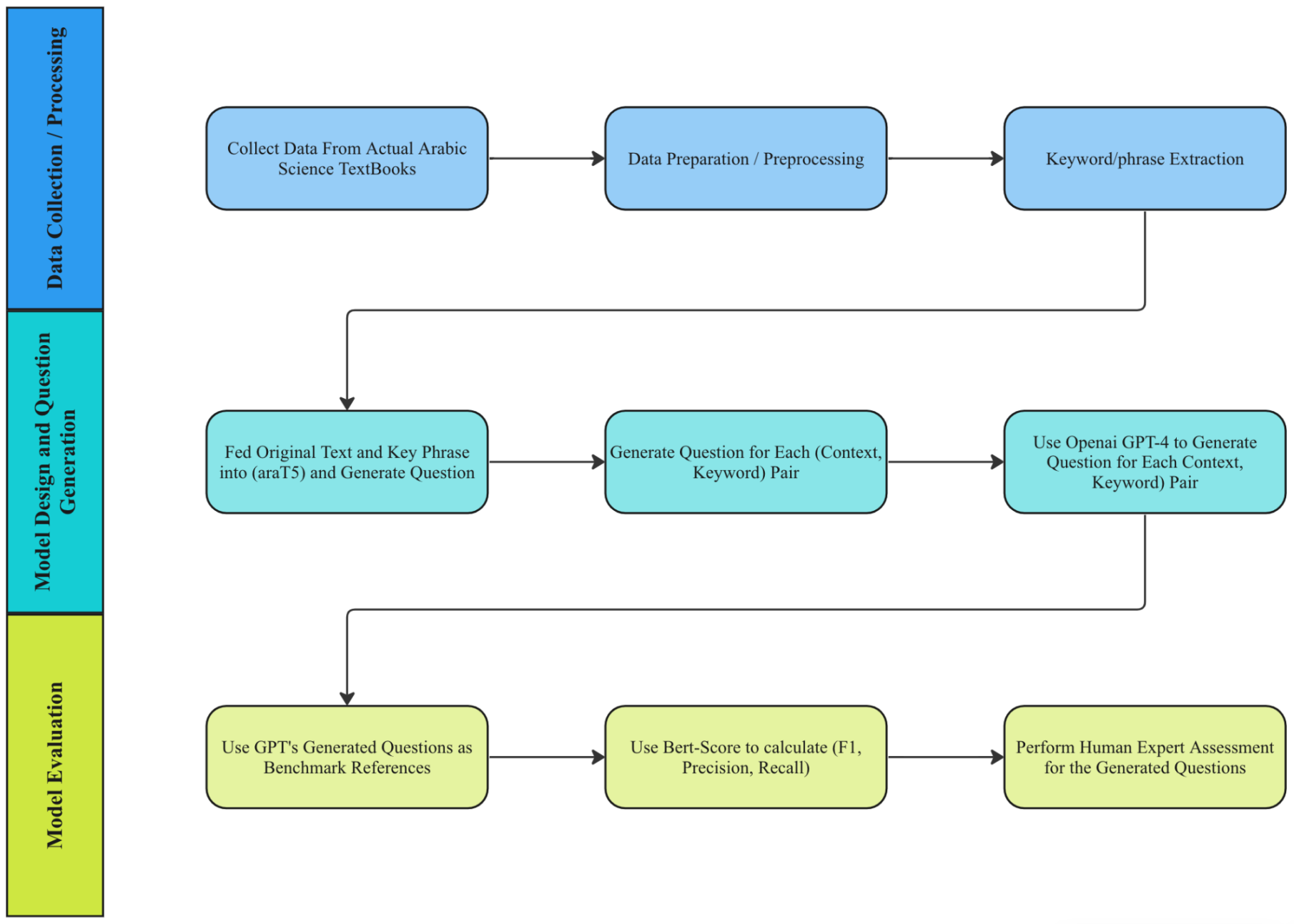

**Fig. 1.** System Methodology Applied in this Study.

## 1.1 **Dataset**

This research employed a dataset derived from Palestinian biology textbooks for grades eleven and twelve. The text was manually extracted and categorized according to topics and sections. Analysis of the dataset's content was conducted using a word cloud, as depicted in Figure 2. Comprising 111 entries, the dataset spans various biology subjects. A basic bag-of-words method revealed that the most common words pertain to blood, cells, DNA, and neurons. The dataset's top 20 words are presented in Figure 3.

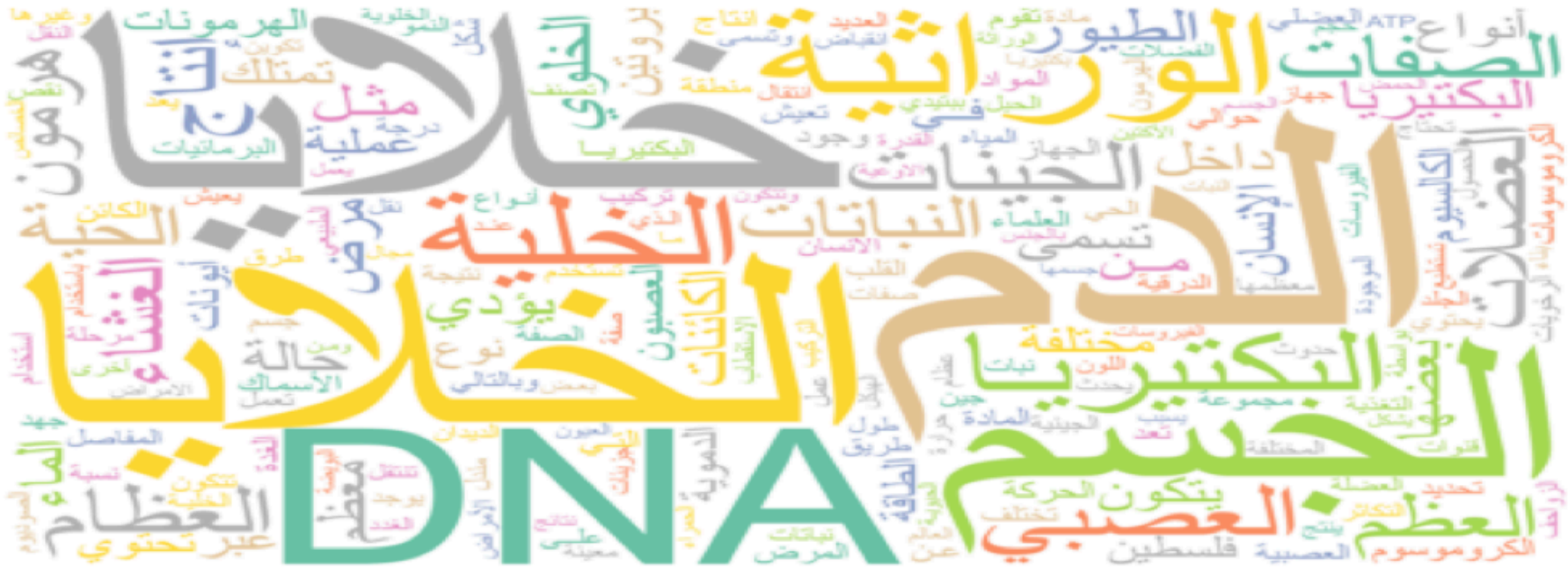

**Fig. 2.** Arabic Dataset Word Cloud.

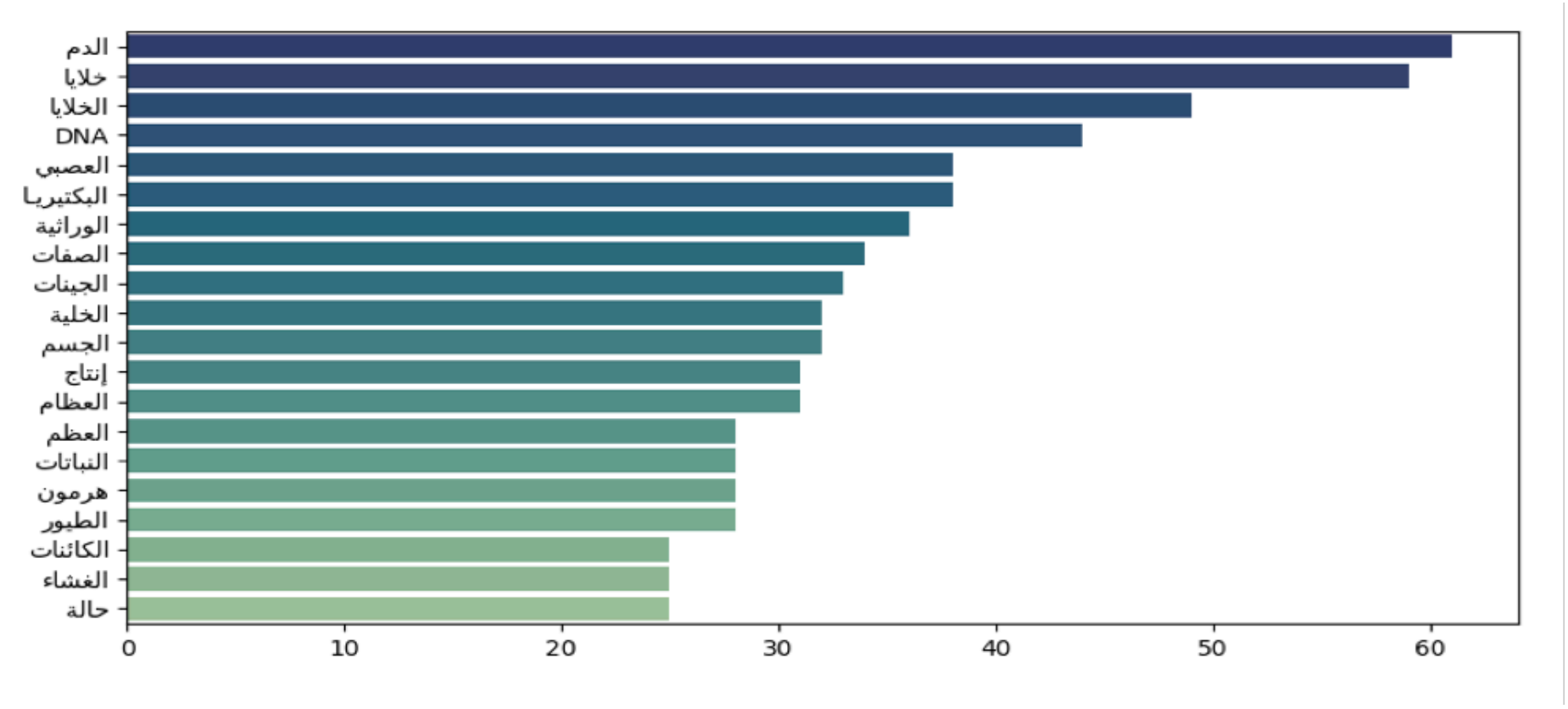

**Fig. 3.** Frequency of Top 20 Words in the Dataset.

## 1.2 Key Phrase Extraction

In our proposed framework, we integrate KeyBert [19], a method that stands out for its simplicity and effectiveness in keyword extraction. KeyBert leverages BERT's sophisticated language models, focusing on the text's context and semantics. Unlike conventional keyword extraction techniques based on frequency, KeyBert embeds the entire text in a high-dimensional vector space to discern the most representative words or phrases of the text's overall theme.

KeyBert functions by first generating document-level representations using BERT embeddings as shown in Figure 4. It then extracts embeddings for N-gram words or phrases as illustrated in Figure 5. Then it applies cosine similarity to identify the words or phrases that are closest to the document's overall vector representation as in Figure 6. Those words or phrases that show the highest similarity are then considered the most descriptive of the document. This method ensures that the keywords extracted are not only contextually right but also encapsulate the essential meaning and theme of the text.

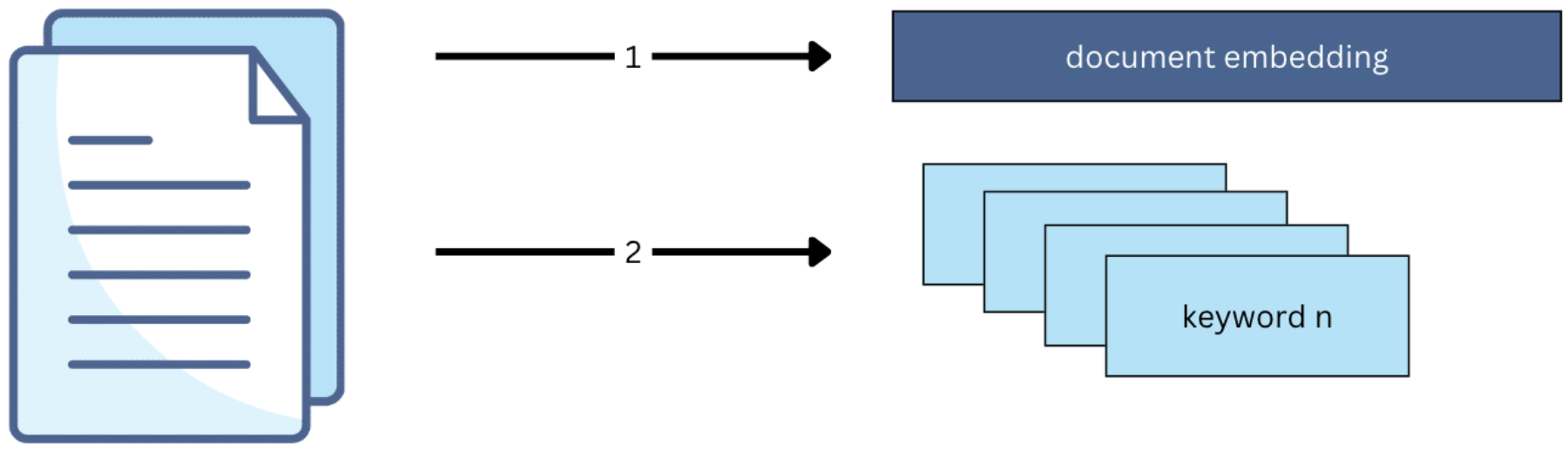

**Fig. 4.** Document-Level Representation Generation with KeyBert.

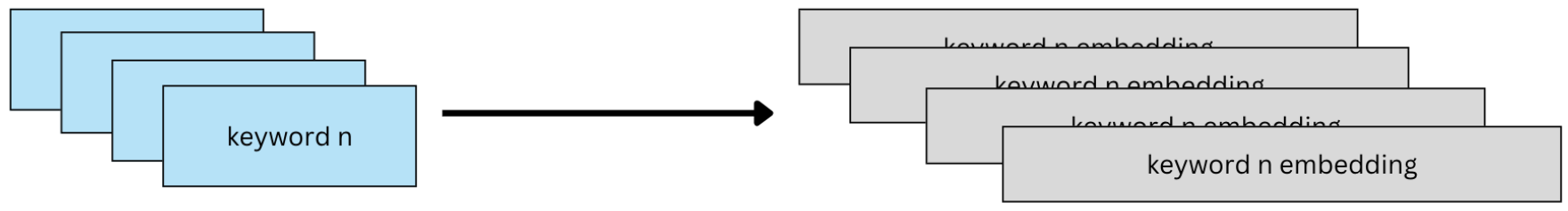

**Fig. 5.** Extracting N-Gram Embeddings with KeyBert.

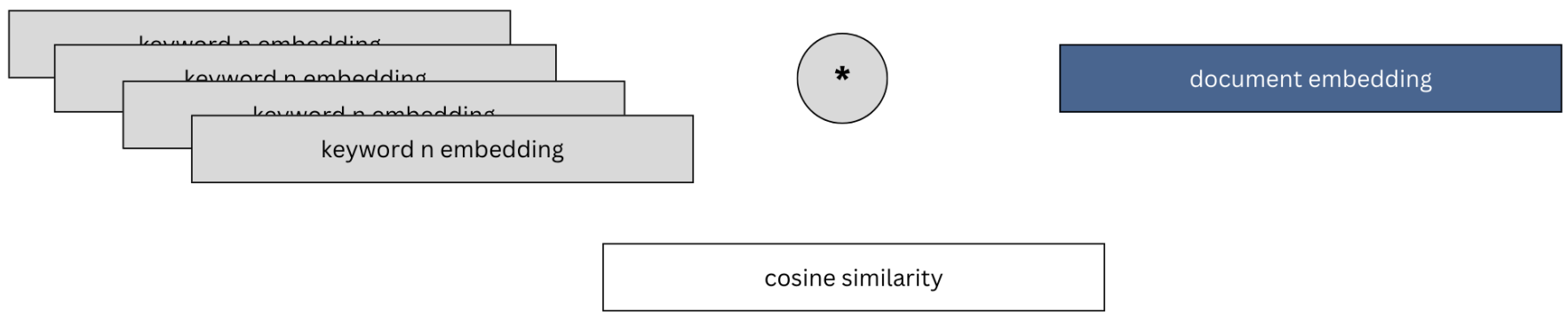

**Fig. 6.** Cosine Similarity Matching in KeyBert.

A primary drawback of the KeyBert-generated phrases is that they can sometimes deviate from the main topic. This occurs because KeyBert relies on basic n-grams to select phrases initially, before applying embedding techniques. Although experimenting with various n-gram sizes might optimize results for a specific dataset, this method is not scalable to other datasets. To address this issue, a more advanced approach, such as incorporating part-of-speech (PoS) tagging, as mentioned in [20], has been integrated into our framework.

This technique involves tagging the original text with PoS tags before calculating embeddings. The KeyphraseVectorizer is comprised of a suite of vectorizers designed to identify keyphrases in text documents using specific part-of-speech patterns. These key phrases are then transformed into a document-keyphrase matrix. This matrix is a structured representation where each row corresponds to a distinct text document and each column represents unique keyphrases. The primary function of this matrix is to track the occurrence frequencies of these keyphrases across the document collection. Advantages of using the KeyphraseVectorizer include the ability to accurately extract keyphrases based on their part-of-speech tags, eliminating the need for predefining n-gram ranges. It also offers the capability to produce document-keyphrase matrices efficiently. The vectorizer supports multiple languages, enhancing its versatility.

The outcome of this process is a set of key phrases that effectively summarize the original texts as shown in Table 1. These phrases are then utilized in the subsequent question-generation step.

## 1.3 ARAT5 Invocation

With the processed text and integrated key phrases, the ARAT5 model is then invoked. This model, pre-trained on extensive Arabic language data, is adept at understanding and manipulating Arabic text. When fed with the prepared input, araT5 leverages its transformer architecture to analyze the context and semantics of the text.

The actual question generation involves the model producing a series of questions based on the input text. The transformer model's capability to understand context allows it to generate questions that are not only syntactically correct but also semantically relevant to the text. The model uses the extracted key phrases from the previous step as potential questionworthy phrases or concepts within the text and formulates questions accordingly as shown in Table 1.

**Table 1.** Arabic Extracted KeyPhrases and Generated Questions Samples.

| Document | Extracted Key Phrase | Generated Question (ARAT5) | Generated Question (GPT4) |
|---|---|---|---|
| يوجد في الطبيعة مجموعة من العناصر يبلغ عددها حوالي 92 عنصرًا، (20 - 25 %) منها عناصر أساسية للكائن الحي ... | عناصر أساسية للكائن الحي | ما هي العناصر الأساسية في الطبيعة؟ | ما هي العناصر الأساسية التي يحتاجها الكائن الحي ليعيش بصحة جيدة؟ |
| يُعد (غريغور مندل) واضع حجر الأساس لعلم الوراثة، وهو أول من توصل إلى نتائج ذاتِ ... | الصفات المندلية لدى الإنسان | ما هي الصفات الوراثية عند الإنسان؟ | ما هي الصفة الوراثية التي كانت تظهر في جميع فراد الجيلين الأول والثاني في تجارب مندل؟ |

### 1.4 **Evaluation**

For a comprehensive evaluation of the questions produced by our proposed framework, we contrast these questions with benchmark questions derived from the same corpus using the state-of-the-art GPT-4. Our assessments indicated that our proposed framework attained a precision of 83.50%, a recall of 78.68%, and an Fl score of 80.95% relative to GPT- 4. Furthermore, a human expert assessed the quality of the generated questions on a scale from 1 to 5, where 1 signifies a very poor question format and 5 represents an outstanding question format. This human evaluation process revealed additional insights into the qualitative aspects of our model's performance. The expert's ratings indicated that most questions scored above 3, suggesting a generally high standard in question formulation. Specifically, the average rating was found to be 4.2 (84%), reflecting the model's ability to formulate coherent and relevant questions. This humancentric evaluation complements the quantitative metrics, offering a more nuanced understanding of the model's capabilities in generating contextually appropriate and linguistically sound questions.

## 3 Results and Discussion

The current study meticulously adopted an exhaustive methodology to guarantee the generation of accurate and robust questions in the Arabic language. This process, as delineated in Figure 1, commenced with the meticulous collection and extraction of data from authentic Arabic textbooks. The extracted texts were then intricately analyzed to extract keywords and key phrases that accurately captured the essence of the original content.

These extracted elements were subsequently employed in conjunction with a pre-trained transformer model, araT5, and Open Al's GPT-4, facilitating the generation of contextually nuanced Arabic questions for each text segment. A comparative analysis involving these questions and those generated by OpenAI offered a multifaceted evaluation of our model's efficacy and efficiency.

Central to our approach was the integration of KeyBert, a sophisticated method utilizing BERT's advanced language models for precise keyword extraction. KeyBert's technique involves creating document-level representations through BERT embeddings and then extracting embeddings for N-gram words or phrases. This method ensures the extraction of keywords that are not only relevant but also deeply representative of the text's context. To enhance the accuracy of KeyBert and mitigate its occasional deviations from the main topic due to its reliance on basic n-grams, our approach included an advanced strategy incorporating Part-of-speech (PoS) tagging. This addition augmented the precision of keyphrase extraction by taking into account the intricate linguistic structure of the text.

This refined process yielded a collection of key phrases that succinctly summarized the original texts, forming the basis for the subsequent step of question generation. Utilizing the araT5 model, which leveraged both the processed text and the integrated key phrases, our system generated a series of questions. These questions were not only syntactically accurate but also deeply aligned with the context and content of the source material, thereby reflecting the model's understanding of the subject matter.

The effectiveness of our model was rigorously evaluated against benchmark questions derived from the same corpus but generated using OpenAI's advanced GPT-4 Turbo. This comparative evaluation employed a range of metrics, revealing a precision rate of 83.50%, a recall rate of 78.68%, and an Fl score of 80.95%. These metrics collectively indicate the high efficiency and accuracy of our framework in generating relevant and contextually appropriate questions as shown in Table 2.

Furthermore, to gain a more nuanced understanding of the model's performance, a human evaluation was conducted. Experts in the field assessed the quality of the generated questions, resulting in an average rating of 4.2 out of 5 (84%). This high rating reflects the model's ability to produce questions that are not only technically sound but also pedagogically valuable, enhancing the learning experience for students.

This comprehensive evaluation, integrating both quantitative and qualitative measures, provided deep insights into the proficiency of our model. It demonstrated the model's capability to generate questions that are coherent, relevant, and aligned with the educational objectives of the Arabic language curriculum. The success of this approach sets a precedent for future research in automated question generation, highlighting the potential of NLP techniques in educational settings. The findings from this study underscore the importance of adapting NLP tools to specific linguistic contexts and suggest possibilities for further development in this domain, including the expansion of methodologies to other languages and subjects, and the incorporation of more sophisticated NLP strategies for even greater precision and relevance in question generation.

**Table 2.** Model Evaluation Metrics and Human Ratings.

| Precision | Recall | F1 Score | Human Evaluation |
|---|---|---|---|
| 83.50% | 78.68% | 80.95% | 84% |

## 4 Conclusion and Future Work

This study marks a significant advancement in automated question generation for Arabic language education, leveraging cutting-edge natural language processing (NLP) techniques. Addressing Arabic's linguistic complexities, our approach integrated KeyBert for keyword extraction and the araT5 transformer model for question generation, achieving high accuracy and efficiency. The model's impressive performance, with a precision of 83.50%, recall of 78.68%, and Fl score of 80.95%, coupled with a human expert rating of 4.2 out of 5, underscores its effectiveness in generating contextually relevant and educationally valuable questions.

Our research highlights the importance of tailoring NLP solutions to specific languages, as evidenced by the comparative analysis with OpenAI's GPT-4. This approach opens up new avenues in educational technology, suggesting the potential applicability of our methodology across various languages and subjects. The success of this project underscores the role of NLP in creating dynamic, adaptive educational tools, and sets the stage for future exploration in NLP and education. The findings suggest promising directions for expanding these methods to other subjects and languages, potentially transforming how educational content is delivered and assessed. This research not only contributes a novel approach for Arabic question generation but also establishes a foundation for future advancements in automated question generation for diverse linguistic contexts.